\documentclass[runningheads]{llncs}
\usepackage{graphicx}

\usepackage{tikz}
\usepackage{comment}
\usepackage{amsmath,amssymb} 
\usepackage{bbm}
\usepackage{booktabs}
\usepackage[colorlinks]{hyperref}
\usepackage{graphicx}
\usepackage{subfig}
\usepackage{bm}
\usepackage{multirow}
\usepackage{cancel}

\usepackage[accsupp]{axessibility}  
\usepackage{algorithm}
\usepackage{algorithmic}

\usepackage{epigraph}

\begin{document}
\pagestyle{headings}
\mainmatter
\def\ECCVSubNumber{100}  

\title{Knowledge Condensation Distillation} 

\titlerunning{Knowledge Condensation Distillation}
%
\author{Chenxin Li\inst{1}
\and
Mingbao Lin\inst{2}\and
Zhiyuan Ding\inst{1}\and
Nie Lin\inst{4}\and
Yihong Zhuang\inst{1} \and
\\ Yue Huang\inst{1,3}\thanks{Corresponding Author} \and
Xinghao Ding \inst{1,3} \and
Liujuan Cao \inst{1}
}
\authorrunning{C. Li et al.}
%
\institute{School of Informatics, Xiamen University \and
Tencent Youtu Lab  \and
Institute of Artificial Intelligence, Xiamen University
\and Hunan University \\
\email{chenxinli@stu.xmu.edu.cn \quad linmb001@outlook.com} \
\email{dingzhiyuan@stu.xmu.edu.cn \quad nielin@hnu.edu.cn}\\  \email{\{zhuangyihong,yhuang2010,dxh,caoliujuan\}@xmu.edu.cn }
}

\maketitle


\begin{abstract}
Knowledge Distillation (KD) transfers the knowledge from a high-capacity teacher network to strengthen a smaller student.
Existing methods focus on excavating the knowledge hints and transferring the whole knowledge to the student.
However, the knowledge redundancy arises since the knowledge shows different values to the student at different learning stages.
In this paper, we propose Knowledge Condensation Distillation (KCD). Specifically, the knowledge value on each sample is dynamically estimated, based on which an Expectation-Maximization (EM) framework is forged to iteratively condense a compact knowledge set from the teacher to guide the student learning. 
Our approach is easy to build on top of the off-the-shelf KD methods, with no extra training parameters and negligible computation overhead.
Thus, it presents one new perspective for KD, in which the student that actively identifies teacher's knowledge in line with its aptitude can learn to learn more effectively and efficiently.
Experiments on standard benchmarks manifest that the proposed KCD can well boost the performance of student model with even higher distillation efficiency. 
Code is available at \url{https://github.com/dzy3/KCD}.

\end{abstract}

\section{Introduction}
\label{sec:intro}


Though deep neural networks (DNNs) have achieved great success in computer vision, most advanced models are too computationally expensive to be deployed on the resource-constrained devices.
To address this, the light-weight DNNs have been explored in the past decades. Typical methods include network pruning~\cite{linHRankFilterPruning2020a}, parameter quantization~\cite{yamamotoLearnableCompandingQuantization2021} and neural architecture search~\cite{caiProxylessNASDirectNeural2019}, \emph{etc}.
Among all these methods, knowledge distillation~\cite{hinton2015distilling} is widely integrated into their learning frameworks, whereby the original cumbersome model (teacher) transfers its knowledge to enhance the recognition capacity of its compressed version, \emph{a.k.a.} student model. 
Due to its flexibility, KD has received ever-increasing popularity in varieties of vision tasks.


\begin{figure}[!t]   
	\centering	   
	\includegraphics[width=0.9\linewidth,keepaspectratio]{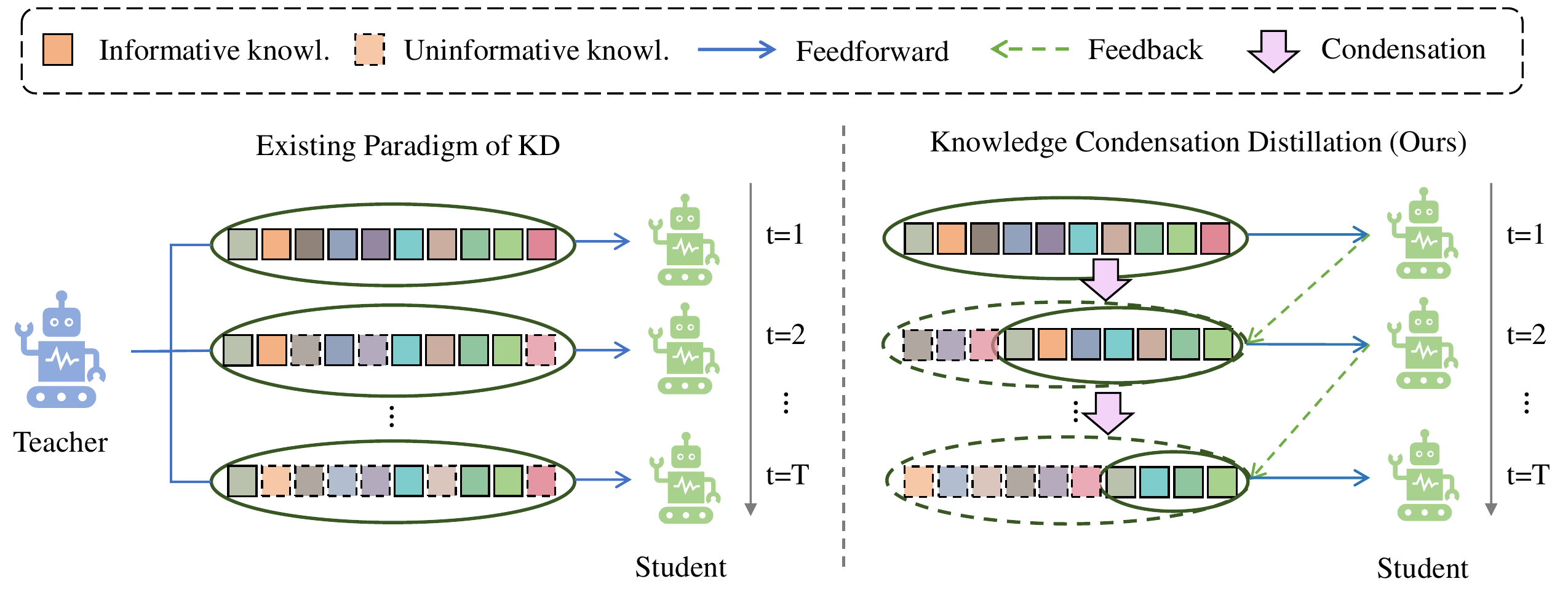}         
	\caption{ 
 Comparison between existing KD paradigm and our KCD. \textbf{Left}: Existing paradigm transfers the complete knowledge points from teacher model across the entire training process, regardless  of  the  varying  values  to  the  student at  different  stages.  \textbf{Right}:  Knowledge points  are  first  estimated  based  on  the  current  capacity  of  the  student, and then condensed to a compact yet informative sub-part for student model.
	}
  \label{fig:mv}    
\end{figure}


In most existing studies on KD \cite{hinton2015distilling,romero2014fitnets,komodakis2017paying,tianContrastiveRepresentationDistillation2020,olvera2010review,chen2021distilling,li2022distilling,li2022learning}, the knowledge hints of the whole sample space, such as soft predictions~\cite{hinton2015distilling}, intermediate representations~\cite{romero2014fitnets}, attention maps~\cite{komodakis2017paying}, \emph{etc}, are transferred to the student model across the entire training process as illustrated in Fig.\,\ref{fig:mv}(\textbf{Left}).
However, these methods neglect the changing capacity of the student model at different learning stages.
Specifically, all the knowledge points from teacher model are informative enough for the student model at its infant learning stage.
However, as the learning proceeds, the value of different knowledge points starts to vary for the student. 
For example, the ``well-memorized'' knowledge points have a relatively limited impact to the student at the later training stages.
Consequently,  the concern regarding redundancy of knowledge transfer arises in existing studies, whereby the student model passively receives all the knowledge points from the teacher.
This further 
poses two severe issues: 
\textbf{(1)} {Training burden}. The redundant knowledge requires not only additional memory storage, but also prolongs the training time. 
\textbf{(2)} {Poor performance}. The redundancy prevents the student model from concentrating enough on the more informative knowledge, which weakens the learning efficacy of the student model.




To overcome the above challenge, as shown in Fig.\,\ref{fig:mv}(\textbf{Right}), this paper presents a new perspective for KD, the cores of which are two main folds: 
\textbf{(1)} A feedback mechanism is introduced to excavate various values of the teacher's knowledge for the student at different training stages.
\textbf{(2)} The student actively identifies the informative knowledge points and progressively condenses a core knowledge set for distillation.
To this end, we propose a Knowledge Condensation Distillation (KCD) paradigm, whereby
the label of knowledge value to student model is encoded as a latent variable, and an Expectation-Maximization (EM) framework is forged to iteratively condense the teacher's knowledge set and distill the student model.
Furthermore, given the local-batch training fashion in student learning, we propose an Online Global Value Estimation (OGVE) module to dynamically estimate the knowledge value over the global knowledge space.
To generate a compact yet effective encoding of teacher's knowledge, %
we develop a Value-Adaptive Knowledge Summary (VAKS) module to adaptively preserve high-value knowledge points, remove the valueless points as well as augment the intermediate ones.
We conduct extensive experiments on two benchmarks, CIFAR100 \cite{CIFAR100} and ImagetNet \cite{IMAGENET}, and many representative settings of teacher-student networks in KD.
We show that our KCD can be well built upon majorities of existing KD pipelines as a plug-and-play solution, without bringing extra training parameters and computation overhead.


Our contributions are summarized as follows:
\begin{itemize}
  \item We propose a novel KD paradigm of knowledge condensation, wherein the knowledge to transfer is actively determined by the student model, and a concise encoding condensed from the whole knowledge set is utilized in KD.

  \item We derive an Expectation-Maximization framework to accomplish our knowledge condensation distillation by iteratively performing knowledge condensation and model distillation.
  
  
  
  \item We propose an OGVE module to acquire an approximate global estimator of knowledge value while utilizing only local training statistics.
  We further present a VAKS module to harmonize the trade-off between compactness and informativeness of knowledge condensation encoding.
   
  
  \end{itemize}


\section{Related Work}

\textbf{Knowledge Distillation}.
The pioneering KD work dates back to~\cite{hinton2015distilling}, where the soft probability distribution from the teacher is distilled to facilitate the student's training.
Since then, abundant developments have been committed to excavating richer knowledge hints, such as intermediate representation~\cite{romero2014fitnets,heoComprehensiveOverhaulFeature2019}, attention maps~\cite{komodakis2017paying}, instance relation~\cite{tungSimilarityPreservingKnowledgeDistillation2019,parkRelationalKnowledgeDistillation2019}, self-supervised embedding~\cite{tianContrastiveRepresentationDistillation2020,xuKnowledgeDistillationMeets2020a} and so on. 
All these methods transfer the knowledge on all training instances to the student regardless of different training stages. Differently, we study the redundancy of the teacher's knowledge and emphasize the significance of making the student model actively condense an efficient knowledge set for learning.

One more recent study~\cite{xu_ComputationEfficientKnowledgeDistillation_2020} considers the efficiency issue of KD by identifying the most informative samples in each training batch. Our method differs from the following aspects. First, the study~\cite{xu_ComputationEfficientKnowledgeDistillation_2020} explores the difference of computation overheads in the forward passes of the teacher and student models, and fixes the knowledge set during the distillation process.
As a comparison in our method, the knowledge set is dynamically condensed and explicitly encodes the patterns of the student model during training.
Second, we estimate the knowledge value over the complete sample space rather than every single batch, which is more accurate and comprehensive.

\textbf{Coreset Construction}.
Another related literature is the problem of coreset construction~\cite{har2007smaller,sener2017active}. 
The main idea behind them is that a learning agent can still perform well with fewer training samples by selecting data itself.
Most existing works~\cite{katharopoulos2018not,wang2018dataset,toneva2018empirical,mirzasoleiman2020coresets,zhang2021efficient} construct this coreset by importance sampling. For example, 
In~\cite{katharopoulos2018not}, sample importance is estimated via the magnitude of its loss gradient \emph{w.r.t.} model parameters.
CRAIG~\cite{mirzasoleiman2020coresets} selects a weighted coreset of training data that closely estimates the full gradient by maximizing a submodular function.
Wang~\emph{et al}.~\cite{wang2018dataset} distilled the knowledge from the entire dataset to generate a synthetic smaller one.
These ideas inspire us to seek a core component of the whole knowledge set from the teacher to realize an efficient KD.

\begin{figure*}[!t]  
	\centering	   
	\includegraphics[width=0.85\textwidth,keepaspectratio]{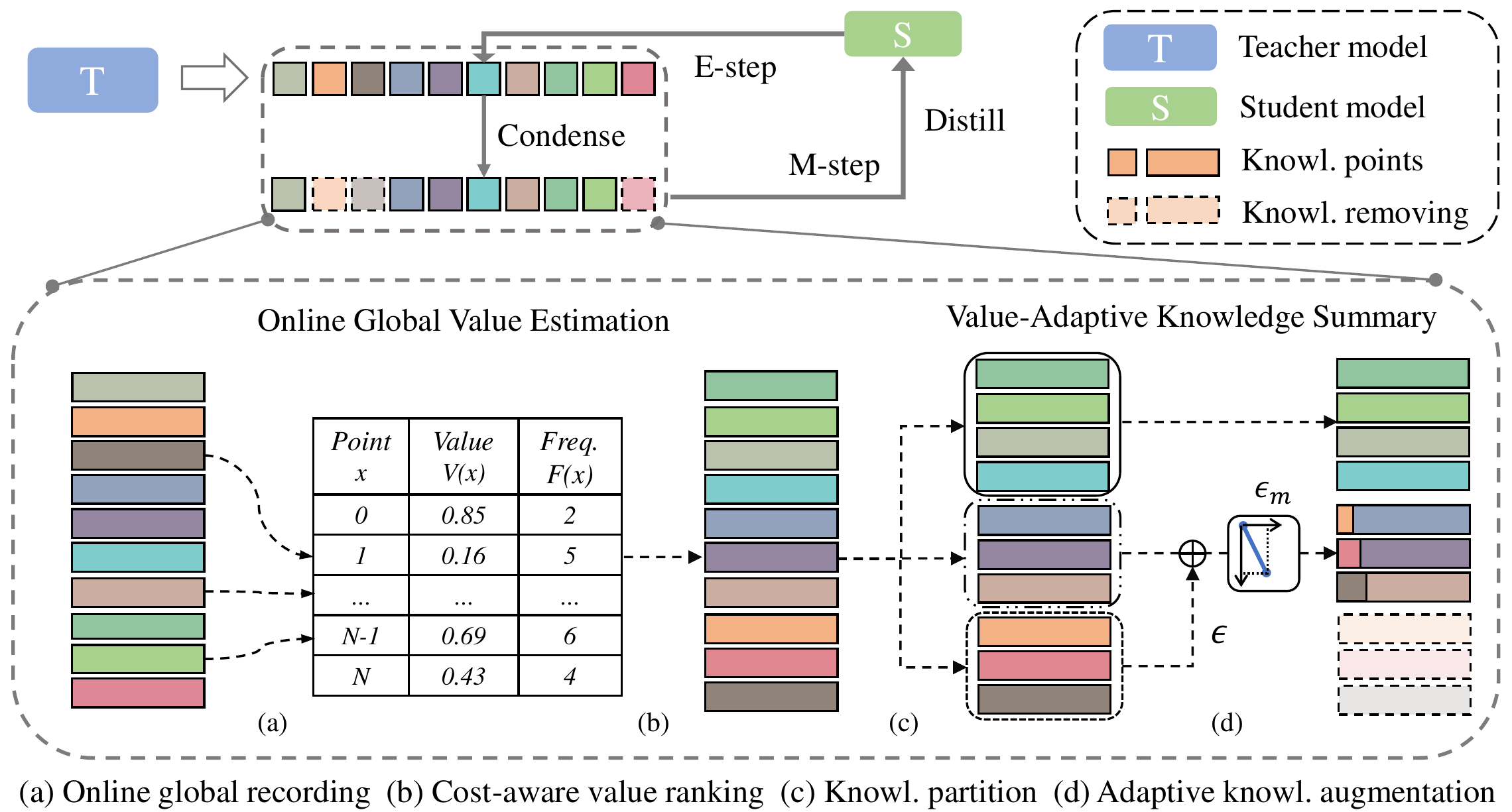}
	\caption{ 
	Overview of the proposed KCD framework. The knowledge condensation and student distillation are optimized iteratively in an EM framework. 
   } \label{fig:pipeline}   
\end{figure*}

\section{Methodology}
\label{method}
\subsection{Preliminaries}
In the task of knowledge distillation (KD), we are given a training dataset $\mathcal{X}$, a pre-trained teacher model $\mathcal{T}$ and a to-be-learned student model $\mathcal{S}$. Hinton~\emph{et al.}~\cite{hinton2015distilling} proposed to minimize the cross-entropy loss between the output probability $p_{\mathcal{T}}(x)$ of the teacher and that $p_{\mathcal{S}}(x)$ of the student:
\begin{equation}
  \label{eq:KD}
  \mathcal{L}_{KD} = -\sum_{x \in \mathcal{X}}p_{\mathcal{T}}(x)\log\big(p_{\mathcal{S}}(x)\big).
\end{equation}

Denoting each pair $\big(x,p_\mathcal{T}(x)\big)$ as a knowledge point, the teacher $\mathcal{T}$ in essence provides a knowledge set $K= \{\big(x,p_{\mathcal{T}}(x)\big)|x \in \mathcal{X}\}$, which is then transferred to the student $\mathcal{S}$. In conventional KD, the knowledge set $K$ is fixed across the whole distillation process, despite different learning stages of the student model. As a core distinction, we propose to transfer simply a concise knowledge encoding $\hat K$ with $|\hat K|<|K|$, where the knowledge points are most valuable and adapt to the demand of the student model at different periods. 

In what follows, we show that the efficient coding $\hat K$ can be deduced by the Expectation-Maximization (EM) framework that encodes the knowledge value for the student model as a latent variable $Y$, with which, we can identify the most valuable components in knowledge set $K$. Fig.\,\ref{fig:pipeline} shows the overview of the proposed method.

\subsection{Knowledge Condensation Distillation}
\label{sec:EM}
The goal of KD in Eq.\,(\ref{eq:KD}) is to learn the parameter $\theta$ of student model in order to maximize the negative cross-entropy between the teacher $\mathcal{T}$ and student $\mathcal{S}$: 
\begin{equation}
  \label{eq:EM1}
  \hat{\theta}=\underset{\theta}{\arg \max }\sum_{x \in \mathcal{X}}\sum_{c \in \mathcal{C}}p_{\mathcal{T}}(x, c) \log p_{\mathcal{S}}(x, c ; \theta),
\end{equation}
where $\mathcal{C}$ denotes the class space.
Instead of transferring the complete knowledge set $K = \{ \big(x, p_{\mathcal{T}}(x)\big)|x\in\mathcal{X} \}$, we introduce a binary value variable $\mathcal{Y} \in \{0, 1\}^{|K|}$, the $i$-th value of which indicates if the $i$-th knowledge point is valuable to the student. In this way, the traditional optimization of Eq.\,(\ref{eq:EM1}) in our setting becomes:
\begin{equation}
  \label{eq:EM2}
  \hat{\theta}=\underset{\theta}{\arg \max }\sum_{x \in \mathcal{X}} \sum_{c \in \mathcal{C}} p_{\mathcal{T}}(x, c) \log \sum_{y \in \mathcal{Y}}{p_{\mathcal{S}}(x, c, y ; \theta)}.
\end{equation}

To maximize this objective, we consider its low-bound surrogate:
\begin{equation}
  \label{eq:EM3}
  \begin{aligned}
  &~~~~~~\sum_{x \in \mathcal{X}} \sum_{c \in \mathcal{C}} p_{\mathcal{T}}(x, c) \log \sum_{y \in \mathcal{Y}}{p_{\mathcal{S}}(x, c, y ; \theta)} \\
  &=\sum_{x\in\mathcal{X}} \sum_{x\in\mathcal{C}} p_{\mathcal{T}}(x, c) \log \sum_{y \in \mathcal{Y}}Q(y)\frac{{p_{\mathcal{S}}(x, c, y ; \theta)}}{Q(y)}\\
  &\geq  \sum_{x \in \mathcal{X}} \sum_{c \in \mathcal{C}} p_{\mathcal{T}}(x, c) \sum_{y \in \mathcal{Y}}Q(y)\log \frac{{p_{\mathcal{S}}(x, c, y ; \theta)}}{Q(y)},
\end{aligned}
\end{equation}
where $Q(y)$ denotes the distribution over the space of value labels $\mathcal{Y}$ so that $\sum_{y \in\mathcal{Y}}Q(y)=1$.
Note that we derive the last step based on Jensen's inequality where the equality holds if and only if $\frac{p_{\mathcal{S}}(x, c, y ; \theta)}{Q(y)}$ is a constant~\cite{li2020prototypical}. 
Under this condition, the distribution $Q(y)$ should be:
\begin{equation}
  \label{eq:EM4}
Q(y) = \frac{p_{\mathcal{S}}(x, c, y ; \theta)}{\sum_{y\in\mathcal{Y}} p_{\mathcal{S}}(x, c, y ; \theta)}= \frac{p_{\mathcal{S}}(x, c, y ; \theta)}{ p_{\mathcal{S}}(x, c; \theta)}=p_{\mathcal{S}}(y;x, c, \theta).
\end{equation}

Removing the constant term $-\sum_{y \in \mathcal{Y}}Q(y)\log Q(y)$ in Eq.\,(\ref{eq:EM3}) and  combining  Eq.\,(\ref{eq:EM4}) lead to our final optimization:
\begin{equation}
  \label{eq:EM5}
  \sum_{x\in \mathcal{X}}\sum_{c \in \mathcal{C}} p_{\mathcal{T}}(x, c)\sum_{y \in \mathcal{Y}}p_{\mathcal{S}}(y;x,c,\theta)\log p_{\mathcal{S}}(x,c,y; \theta).
\end{equation}

The maximization of the above problem can be realized by Expectation-Maximization (EM) algorithm, as elaborated below:

\textbf{E-step}.
%
In this step, we aim to evaluate value distribution $Q(y)=p_{\mathcal{S}}(y;x,c, \theta)$. Before that, we first discuss how to measure the value of each knowledge point $(x, p_{\mathcal{T}(x)})$, insight of which is two-fold: First, it has been verified that the average prediction entropy loss decreases drastically if a model is distilled by knowledge hints from a teacher model, instead of being trained solely~\cite{muller_WhenDoesLabel_2020}. This reflects the contribution of knowledge points to the training of the student model. Second, as discussed in~\cite{shen_LabelSmoothingTruly_2021}, the knowledge which encodes informative semantic structure tends to require more training time for the student model to fit well.

These two insights indicate that the prediction entropy loss can be an option to measure the knowledge value. Besides, informative knowledge tends to cause a larger entropy loss.
Therefore, given a knowledge point $(x, p_{\mathcal{T}(x)})$, we utilize its prediction entropy to measure its value:
\begin{equation}
  \label{eq:entropy}
  V(x)=-\sum_{c\in\mathcal{C}} {p_\mathcal{S}(x,c)}\log {p_\mathcal{S}(x,c)}.
\end{equation}

With the prediction entropy, in order to estimate $p_{\mathcal{S}}(y;x,c, \theta)$, we further conduct a ranking operation in an decreasing order \emph{w.r.t.} $V(x)$ over $\mathcal{X}$. Then, based on the ranking position $\mathcal{R}_V(x) \in \{0,1,\cdots,N \}$, we derive the relative likelihood probability about knowledge value:
\begin{equation}
  \label{eq:prob}
  p_{\mathcal{R}_V}(y;x,\theta)=1-\frac{\mathcal{R}_V(x)}{|\mathcal{X}|}.
\end{equation}

Then the likelihood of value label $p_{\mathcal{S}}(y;x,c, \theta)$ can be determined by a threshold $\tau$: 
$p_{\mathcal{S}}(y;x,c, \theta) = 1$ if $p_{\mathcal{R}_V}(y;x,\theta)\geq \tau$, and 0 otherwise.

\textbf{M-step}.
%
After E-step, the optimized object of Eq.\,(\ref{eq:EM5}) can be re-written as:
\begin{equation}
  \label{eq:EM6}
  \begin{aligned}
  &~~~~\sum_{x \in \mathcal{X}}\sum_{c \in \mathcal{C}} p_{\mathcal{T}}(x, c)\sum_{y \in \mathcal{Y}}p_{\mathcal{S}}(y;x,c, \theta)\log p_{\mathcal{S}}(x,c,y; \theta)\\
  &=\sum_{x\in \mathcal{X}}\sum_{c \in \mathcal{C}} p_{\mathcal{T}}(x, c)\sum_{y \in \mathcal{Y}}\mathbbm{I}(p_{\mathcal{R}_V}(y;x,\theta)\geq \tau) \log p_{\mathcal{S}}(x,c,y; \theta),
  \end{aligned}
\end{equation}
where $\mathbbm{I}(\cdot)$ returns 1 if its input is true, and 0 otherwise. When the training samples are not provided, we assume a uniform priori over $y$ (0 or 1):
\begin{equation}
  \label{eq:EM7}
p_{\mathcal{S}}(x, c, y ; \theta) =  p_{\mathcal{S}}(x, c; y , \theta) p_{\mathcal{S}}(y; \theta)= \frac{1}{2} \cdot p_{\mathcal{S}}(x, c; y , \theta),
\end{equation}
where $p_{\mathcal{S}}(y; \theta) = \frac{1}{2}$ due to the premise of uniform distribution.
Then the distribution $p_{\mathcal{S}}(x, c; y , \theta)$ is only conditioned on the estimated value label $Y$, \emph{i.e.}, $\mathbbm{I}(p_{\mathcal{R}_V}(y;x,\theta)\geq \tau)$. We conduct distillation only upon the knowledge points with label $y=1$. Thus, we can re-write the maximum estimation in Eq.\,(\ref{eq:EM1}) as:
\begin{equation}
  \label{eq:EM8}
  \hat\theta=\underset{\theta}{\arg \min }\sum_{ x\in\mathcal{X}|Y(x)=1 }\sum_{c \in \mathcal{C}}-p_{\mathcal{T}}(x, c) \log p_{\mathcal{S}}(x, c ; \theta),
\end{equation}
where $x\in\mathcal{X}$ can be used for distillation only when the condition $y=1$.

Consequently, our KCD iteratively performs E-step and M-step. The former aims to find the distribution of label $Y$ and the concise knowledge encoding $\hat{K}$ comprises these knowledge points with $y=1$, while the latter implements efficient distillation upon the concise set $\hat{K}$.
However, current neural networks are trained in a batch fashion where a small portion of samples are fed forward each time. These local sample batches barricade a direct extraction of the concise knowledge set $\hat{K}$ from the whole training dataset $\mathcal{X}$. In what follows, we further propose an online global value estimation to solve this problem.

\subsection{Online Global Value Estimation}
\label{sec:OGVE}
In order to condense a valuable knowledge set $\hat{K}$ in a global fashion, we design an online global value estimation (OGVE) to derive the global statistics of the whole training dataset $\mathcal{X}$ which consists of an online value recording and cost-aware ranking below:


\textbf{Online Value Recording}. The estimation of $p_{\mathcal{S}}(y;x,c, \theta)$ is conducted by 
$p_{\mathcal{R}_V}(y;x,\theta)$ over the whole (global) sample space $\mathcal{X}$ at the E-step.
However, only a small-portion sub-set (local) of knowledge can be available at each training iteration. Besides, the same sample $x$ might appear frequently at different training stages.
To alleviate this issue, we propose to consider the historical statistics of $x$.
Technically, when $x$ is fed to the network at a particular training iteration, we first count the frequency of $x$ ever involving in the training, denoted as $F(x)$. Also, we calculate its prediction entropy $V(x)$ at current training iteration using Eq.\,(\ref{eq:entropy}). Then, the global value of a knowledge point $\big(x, p_{\mathcal{T}(x)}\big)$ is updated in an online moving-average fashion as:
\begin{equation}
  \label{eq:ma}
 V^{F(x)}(x) = \frac{F(x)-1 }{F(x)}\times  V^{F(x)-1} + \frac{1}{F(x)}V(x).
\end{equation}


%
\textbf{Cost-Aware Ranking}.
Based on the recorded global statistics of $V^{F(x)}(x)$, we can obtain a more accurate ranking of $R_V(x)$ without introducing any additional overhead.
However, in the current design, the ranking order of two knowledge points with a similar value might be the same even though their training  frequencies are very different, which is counter-intuitive, given the fact that the neural network tends to memorize and gives low prediction entropy to these samples that have ever seen more times~\cite{han2018co}. 
Therefore, the knowledge points with similar $ V(x)$ but a higher training cost $F(x)$ should be more critical for the student model and assigned with a top ranking. Considering this, in the ranking operation, we re-weight $ V(x)$ using the training frequency $F(x)$ as:
\begin{equation}
  \label{eq:rank}
  \mathcal{R}_V(x)=\mathop{\text{arg sort}}\limits_{x \in \mathcal{X}} ~ V^{F(x)}(x)\times \big(F(x)\big)^\alpha,
\end{equation}
where $\alpha$ controls the weighted effect of $F(x)$. Eq.\,(\ref{eq:rank}) considers not only the status of $ V(x)$, but also the cost $F(x)$ to achieve this status. 
%


%
Combining $\mathcal{R}_V(x)$ in Eq.\,(\ref{eq:rank}) and $p_{\mathcal{R}_V}(y;x,\theta)$ in Eq.\,(\ref{eq:prob}), we can estimate the value label $Y$. Accordingly, the concise knowledge encoding $\hat{K}$ consists of the knowledge points with $y = 1$.
As the training proceeds, many well learned knowledge points becomes less valuable to the student model. However, the relative likelihood probability $p_{\mathcal{R}_V}(y;x,\theta) \neq 0$ indicates a possibility to be selected again.
%
%
Instead, we further propose a value-adaptive knowledge summary by solving this issue in a divide-and-conquer manner.


\subsection{Value-Adaptive Knowledge Summary} 
\label{sec:VAKS}
Our value-adaptive knowledge summary (VAKS) performs concise knowledge encoding in a two-step fashion including a knowledge partition and an adaptive knowledge augmentation. 
%

\textbf{Knowledge Partition}.
According to our OGVE, we can obtain an explicit label set $Y$. Then, the original knowledge set can be divided into $K_1$ with $y=1$ and $K_0$ with $y=0$. For knowledge points in $K_0$, they are deemed to be valueless thus we choose to directly discard them.
As for $K_1$, based on the relative likelihood probability $\mathcal{R}_V(x)$, we further partition it into a set $K_{1H}$, element of which has a relatively high $\mathcal{R}_V(x)$, and a set  $K_{1L}$, element of which has relatively low $\mathcal{R}_V(x)$, as shown in Fig.\,\ref{fig:pipeline}. Besides, our partition also requires $K_{1L}$ to be in the same size with $K_0$, \emph{i.e.}, $|K_{1L}| = |K_0|$, reason of which will be given in the following adaptive knowledge augmentation.

The knowledge points in $K_{1H}$ are of high possibility to be valuable for the student, thus they can be safely transferred to the student as the conventional KD does.
However, the knowledge in $K_{1L}$ falls into a ``boundary status''. Though considered valuable, they are prone to being less valuable than knowledge in  $K_{1H}$ and easily absorbed by the student. This motivates us to enhance the knowledge points in $K_{1L}$.
One straightforward approach is to introduce the gradient-based distillation \cite{wang2018dataset,yin2020dreaming,zhao2020dataset} to generate new knowledge contents. However, the heavy time consumption barricades its application.
In what follows, we introduce an adaptive knowledge augmentation to reach this goal in a training-free fashion.

\textbf{Adaptive Knowledge Augmentation}.
Our insight of knowledge augmentation comes from the field of adversarial examples~\cite{szegedy2013intriguing,goodfellow2014explaining}, where a subtle perturbation can greatly confuse the model recognition.
Likewise, we also seek a perturbation on the knowledge points in $K_{1L}$. 
It is noteworthy that rather than to find the most disruptive disturbance in adversarial examples, our goal is to use some knowledge-wise perturbation to augment the knowledge points, making them more informative for the student model.

Concretely, denoting $S = \{|K_1|,|K_1|-1,...,|K_{1L}|\}$,
as shown in Fig.\,\ref{fig:pipeline}, we propose to make full use of the removed valueless knowledge in $K_0$ to augment knowledge points in $K_{1L}$ with a very small perturbation ratio $\epsilon $.  as:
\begin{equation}
   \label{eq:aug}
K_{Aug} = \text{Ordered}(K_{1L}) \oplus \text{Ordered}(K_0)\otimes \epsilon(S),   
\end{equation}
where $\text{Ordered}(\cdot)$ reorders its input set in descending according to the value of knowledge point, and $\oplus$ denotes the element-wise adding operation. Recall that $|K_{1L}| = |K_0|$ in our setting, thus, the $\oplus$ is applicable. $\epsilon(\cdot)$ is defined as:
\begin{equation}
  \label{eq:linear}
\epsilon(x') = \frac{\epsilon_m}{|K_{0}|}(x'-|K_1|)+\epsilon_m.
\end{equation}

Thus $\epsilon(S)$ is a set linearly increasing from 0 to a pre-given $\epsilon_{m}$ (see Fig.\,\ref{fig:pipeline}). The intuition of $\epsilon(x')$ is to make the knowledge points with lower-ranking positions \emph{w.r.t.} knowledge value to get more augmentation effect while the ones with higher positions to maintain more their original knowledge contents.

Finally, we obtain the knowledge condensation $\hat{K} = K_{1H} \cup K_{Aug}$.

\begin{algorithm}[!t]
  \label{alg}
  \caption{Knowledge Condensation Distillation}
  \label{alg}
  \textbf{Input}: Training dataset $\mathcal{X}$; 
  a student model $\mathcal{S}$ with learnable parameters $\theta$;
  a full knowledge set $K$ generated by a pre-trained teacher model $\mathcal{T}$.\\
  \textbf{Required}: 
  Number of epochs in a learning stage $T$;
  Desired final knowledge condensation ratio $\rho$.\\
  \textbf{Output}: Distilled student model with parameter $\hat{\theta}$; condensed knowledge encoding $\hat{K}$ ($|\hat{K}|=|K|\cdot \rho$).
  \begin{algorithmic}[1] 
  \STATE Init. $\hat K = K$;
  \FOR{$i =0,...,I$ ~\textbf{epoch} } 
  \STATE \textcolor[rgb]{0.65,0.65,0.65}{\# M-step: Knowledge distillation}\\
    \STATE Distill $\hat\theta$ of student $\mathcal{S}$ on the condensed knowledge $\hat{K}$  via Eq.\,(\ref{eq:EM8});\\
   \STATE \textcolor[rgb]{0.65,0.65,0.65}{\#E-step: Knowledge condensation}\\
  \STATE \textcolor[rgb]{0.65,0.65,0.65}{\#\# Estimate knowledge value over  $K$ via proposed OGIE (Sec.\,\ref{sec:OGVE}) }
  \STATE Cal. knowledge value $V(x)$ over compact (local) knowledge space $\hat K$ via Eq.\,(\ref{eq:entropy}); 
  online update historical recording $V^{F(x)}(x)$ over complete (global) knowledge space $K$ via Eq.\,(\ref{eq:ma});\\
  \IF {$i$ \% $T=0$}
  \STATE Cal. ranking position of knowledge value $\mathcal{R}_V(x)$ via Eq.\,(\ref{eq:rank}); cal. ranking-based likelihood probability $p_{\mathcal{R}_V} (x)$ via Eq.\,(\ref{eq:prob}); \\
  \STATE Binarize $p_{\mathcal{R}_V}(y;x)$ with the threshold $\tau^t$ at current stage $t=i/T$; determine value label $Y$ ($y=1$ or $0$) over complete knowledge space $K$;
  \STATE \textcolor[rgb]{0.65,0.65,0.65}{\#\# Summarize knowledge encoding $\hat K$ via proposed VAKS (Sec.\,\ref{sec:VAKS})};
  \STATE Partition $K$ into $K_1$ and $K_0$ via label $Y$; partition $K_1$ into $K_{1H}$ and $K_{1L}$, \emph{s.t.} $|K_{1L}|=|K_0|$; \\
  \STATE Augment $K_{1L}$ via Eq.\,(\ref{eq:aug}) and Eq.\,(\ref{eq:linear}); \\
  \STATE Summarize compact knowledge encoding via $\hat{K} = K_{1H} \cup K_{Aug}$. \\
  \ENDIF

  
  \ENDFOR

  \end{algorithmic}
  \end{algorithm}

\subsection{Overall Procedure}
\label{sec:efficiency}
The overall procedure of our proposed KCD is depicted in Alg.\,\ref{alg}.
The proposed framework iteratively performs knowledge condensation in E-step and knowledge distillation in M-step, which 
can be practically formulated as a stage-based learning framework.
The total $I$ training epochs are equally divided into $I/T$ learning stages, each with $T$ epochs.
Within each stage, the distillation is conducted for $T$ epochs on the fixed knowledge set, followed by the recording of knowledge value in every training batch (Eq.\,(\ref{eq:ma})).
At the end of each stage, we perform a ranking step across the whole knowledge set \textit{w.r.t.} knowledge value (Eq.\,(\ref{eq:rank})) and knowledge summary (Eq.\,(\ref{eq:aug})). to condense a smaller informative knowledge set. 
The condensed one is then used in the next stage.
%

It is noteworthy that the reduction of computation overhead mainly comes from using more compact knowledge encoding $\hat K$ during KD.
To quantitatively portray this,
absolute cost $C_a$ is calculated by the number of knowledge points used, \textit{e.g.}, $C_a=|K|\cdot I$ for conventional KD.
We further calculate the relative cost $C$ as the rate of $C_a$ between our KCD and the conventional KD baseline:
\begin{equation}
  \label{eq:cost}
  C = \frac{|K|\cdot(\tau^0 + \tau^1 + \cdots + \tau^t + \cdots + \tau^{I/T} )\cdot T}{|K|\cdot I} 
\end{equation}
where $\tau^t$ denotes the threshold of the ranking-based probability $p_{\mathcal{R}_V}$ (Eq.\,(\ref{eq:prob})) for the value label $Y$  
at the $t$-th stage. 
It controls the condensation ratio, 
as $|\hat{K}|=|K|\cdot\tau^t$ at $t$-th stage and
the final condensation rate $\rho=\tau^{I/T}$.
%

\section{Experiments}
\textbf{Datasets}. 
We conduct experiments on two benchmark datasets for KD, namely CIFAR100~\cite{CIFAR100} and ImageNet~\cite{IMAGENET}.
CIFAR100 contains 50K training images with 500 images per class and 10K test images with 100 images per class. The image size is 32$\times$32.
ImageNet is a large-scale classification dataset, containing 1.2 million images over 1K classes for training and 50K for validation. The image size is 224$\times$224.


\textbf{Implementation Details}. 
Following the common practice in \cite{tianContrastiveRepresentationDistillation2020,xuKnowledgeDistillationMeets2020a},
we adopt the stochastic gradient descent (SGD) optimizer with a momentum of 0.9, weight decay of $5\times10^{-4}$.
Batch size is set as 64 for CIFAR-100 and 256 for ImageNet.
For CIFAR100 \cite{CIFAR100}, the learning rate is initialized as 0.05, and decayed by 0.1 every 30 epochs after the
first 150 epochs until the last 240 epochs. 
For ImageNet \cite{IMAGENET}, the learning rate is initialized as 0.1, and decayed by 0.1 every 30 epochs.
Without specification, the hyper-parameters in Alg.\,\ref{alg} is set as follows:
We set $I=240$, $T=40$ for CIFAR100 and $I=90$, $T=15$ for ImageNet. 
We set final condensation rate $\rho=0.7$. The intermediate value of condensation threshold $\tau$ is set as exponential decay after every learning stage, with the initial value of $\tau^0=\sqrt[T/E]{\rho}=0.9423$.
We set $\alpha=0.03$ in Eq.\,(\ref{eq:rank}), and the perturbation rate $\epsilon$ in Eq.\,(\ref{eq:linear}) as
linear growth from 0 to 0.3 ($\epsilon_m=0.3$).

\begin{table*}[!t]
  \centering
  \caption{Test Acc. (\%) of the student networks on CIFAR100. 
  \textbf{Bold} and \underline{underline} denote the best and second best results. The comparison of whether to equip modern methods with our KCD is provided as (+/-). Same-architecture and cross-architecture experiments are shown in two groups of columns.
  }
    \resizebox{1.0\linewidth}{!}{
    \setlength{\parindent}{0pt}
    \setlength{\tabcolsep}{1.6pt}{
    \begin{tabular}{c|ccccc|cccccc}
    \toprule
    \multicolumn{1}{c|}{Teacher} & {W40-2} & {W40-2} & {R56} & {R32x4} & {V13} 
    & 
    {V13} &  {R50} &  {R50} &  {R32x4} &  {R32x4} &  {W40-2}\\
    \multicolumn{1}{c|}{Student} & {W16-2}  & {W40-1} & {R20} & {R8x4} & {V8} 
    & 
    {MN2} &  {MN2} &  {V8}  &  {SN1} &  {SN2} &  {SN1}\\
    \midrule
    Teacher & 75.61  & 75.61  & 72.34  & 79.42  & 74.64 & 74.64  & 79.34  & 79.34  & 79.42  & 79.42  & 75.61 \\
    Student & 73.26  & 73.26  & 69.06  & 72.50  & 70.36 & 64.60  & 64.60  & 70.36  & 70.50  & 71.82  & 70.50 \\
    \midrule
    KD \cite{hinton2015distilling}  & 74.92  & 73.54  & 70.66  & 73.33  & 72.98  & 67.37 & 67.35 & 73.81  & 74.07  & 74.45  & 74.83\\
    FitNet \cite{romero2014fitnets} & 73.58  & 72.24  & 69.21  & 73.50  & 71.02  & 64.14  & 63.16  & 70.69  & 73.59  & 73.54  & 73.73 \\
    AT \cite{komodakis2017paying}  & 74.08  & 72.77  & 70.55  & 73.44  & 71.43  & 59.40  & 58.58  & 71.84  & 71.73  & 72.73  & 73.32\\
    SP \cite{tungSimilarityPreservingKnowledgeDistillation2019}    & 73.83  & 72.43  & 69.67  & 72.94  & 72.68  & 66.30  & 68.08  & 73.34  & 73.48  & 74.56  & 74.52\\
    VID \cite{ahnVariationalInformationDistillation2019a}  & 74.11  & 73.30  & 70.38  & 73.09  & 71.23  & 65.56  & 67.57  & 70.30  & 73.38  & 73.40  & 73.61\\
    RKD \cite{parkRelationalKnowledgeDistillation2019}  & 73.35  & 72.22  & 69.61  & 71.90  & 71.48  & 64.52  & 64.43  & 71.50  & 72.28  & 73.21  & 72.21\\
    PKT \cite{passalisLearningDeepRepresentations2018}  & 74.54  & 73.45  & 70.34  & 73.64  & 72.88  & 67.13  & 66.52  & 73.01  & 74.10  & 74.69  & 73.89\\
    CRD \cite{tianContrastiveRepresentationDistillation2020}  & 75.64 & 74.38  & 71.63  & 75.46  & 74.29 & 69.94 & 69.54  & 74.58  & 75.12  & 76.05  & 76.27\\
    WCoRD \cite{chen2021wasserstein} & 76.11 & 74.72 & \underline{71.92} &  \underline{76.15} & 74.72 & 70.02 & 70.12 & 74.68 & 75.77 & 76.48 & 76.68 \\
    ReviewKD \cite{chen2021distilling} & \underline{76.12} & 75.09 & 71.89 &  75.63 & 74.84 &70.37 & 69.89 & - & 77.45 & 77.78 & \underline{77.14}\\
    SSKD \cite{xuKnowledgeDistillationMeets2020a} & 75.66 & \underline{75.27} & 70.96 & 75.80 & \textbf{75.12} & \underline{70.92} & \underline{71.14} & \textbf{75.72} & \underline{77.91} & \underline{78.37} & 76.92 \\ 

    \midrule
    \multicolumn{1}{c|}{\multirow{2}[0]{*}{KC-KD}}  & 75.70 & 73.84 & 70.75 & 74.05 & 73.44 & 68.61 & 67.94 & 74.41 & 74.33 & 75.19 & 75.60 \\
    & \footnotesize{(+0.78)}  &  \footnotesize{(+0.30)}  &  \footnotesize{(+0.09)}  &  \footnotesize{(+0.72)}  & \footnotesize{(+0.46)}  & \footnotesize{(+1.24)}  & \footnotesize{(+0.59)}  & \footnotesize{(+0.60)}  & \footnotesize{(+0.26)}  & \footnotesize{(+0.74)}  & \footnotesize{(+0.77)}
    \\
    \multicolumn{1}{c|}{\multirow{2}[0]{*}{KC-PKT}} & 75.01 & 74.12 & 72.08 & 74.45 &72.82 & 67.99 & 67.92 & 73.32 &74.60 & 75.79 &75.78\\ 
    &\footnotesize{(+0.47)} & \footnotesize{(+0.67)} & \footnotesize{(+1.74)} & \footnotesize{(+0.81)} & \footnotesize{(-0.06)} & \footnotesize{(+0.86)} & \footnotesize{(+1.40)} & \footnotesize{(+0.31)} & \footnotesize{(+0.50)} & \footnotesize{(+1.10)} & \footnotesize{(+1.89)}
    \\
    \multicolumn{1}{c|}{\multirow{2}[0]{*}{KC-CRD}}  & 75.93 & 74.60 & \textbf{72.11} & 75.78 & 74.38 &69.90 &69.82 & 74.49 & 75.74 & 76.44 & 76.4
    \\
    & \footnotesize{(+0.29)} & \footnotesize{(+0.22)} &\footnotesize{(+0.48)} & \footnotesize{(+0.32)} & \footnotesize{(+0.09)} 
    & \footnotesize{(-0.04)} & \footnotesize{(+0.28)} & \footnotesize{(-0.09)} & \footnotesize{(+0.62)} & \footnotesize{(+0.39)} & \footnotesize{(+0.13)}
    \\
    \multicolumn{1}{c|}{\multirow{2}[0]{*}{KC-SSKD}} & \textbf{76.24} & \textbf{75.35}  & 71.31 & \textbf{76.48}  & \underline{74.93}
    & \textbf{71.32} & \textbf{71.29} &\underline{75.65} & \textbf{78.28} & \textbf{78.59} & \textbf{77.61}
    \\
    & \footnotesize{(+0.58)} & \footnotesize{(+0.08)} & \footnotesize{(+0.35)} & \footnotesize{(+0.68)} & \footnotesize{(-0.21)} 
    & \footnotesize{(+0.40)} & \footnotesize{(+0.15)} & \footnotesize{(-0.07)} & \footnotesize{(+0.37)} & \footnotesize{(+0.22)} & \footnotesize{(+0.69)}
    \\
    \bottomrule
    \end{tabular}
    }%
    }
  \label{tab:SOTA}%
\end{table*}%

\begin{table}[!t]
  \centering
  \caption{
   Test Acc. (\%) with computation cost $C$ on CIFAR100, compared with the only existing method UNIX~\cite{xu_ComputationEfficientKnowledgeDistillation_2020} that focuses on distillation efficiency.
  }
  \resizebox{\linewidth}{!}{
    \setlength{\parindent}{0pt}
    \setlength{\tabcolsep}{1.8pt}{
    \begin{tabular}{ccccccccc}
    \toprule
    Teacher & WRN-40-2 & WRN-40-2 & resnet56 & VGG13 & VGG13 & ResNet50 \\
    Student & WRN-16-2 & WRN-40-1 & resnet20 & VGG8 & MobileNetV2 & VGG8 \\
    \midrule
    KD    & 74.92 (100\%) & 73.54 (100\%) & 70.66 (100\%) & 72.98 (100\%) & 67.37 (100\%) & 73.81 (100\%) \\
    UNIX-KD & 75.19 (75.3\%) & 73.51 (73.1\%) & 70.06 (76.0\%) & 73.18 (76.4\%) & 68.47 (77.5\%) & 73.62 (68.9\%) \\
    UNIX-KD\dag & 75.25 (81.6\%)  & \textbf{74.18} (81.6\%)) & 70.19 (81.6\%)  & 73.27 (81.6\%)  & 68.58 (81.6\%)  & 74.24 (81.6\%)  \\  
    KC-KD & \textbf{75.70} (81.6\%) & 73.84 (81.6\%) & \textbf{70.75} (81.6\%)  & \textbf{73.44} (81.6\%)  & \textbf{68.61} (81.6\%)  & \textbf{74.41} (81.6\%)  \\
    \bottomrule
    \end{tabular}%
  }
  }
  \label{tab:UNIX}%
\end{table}%

\subsection{Comparisons with State-of-the Arts}
\textbf{Results on CIFAR100}.
We make comparison to various representative state-of-the-art KD methods, including vanilla KD~\cite{hinton2015distilling}, FitNet~\cite{romero2014fitnets}, AT~\cite{komodakis2017paying}, SP~\cite{tungSimilarityPreservingKnowledgeDistillation2019}, VID~\cite{ahnVariationalInformationDistillation2019a}, RKD~\cite{parkRelationalKnowledgeDistillation2019}, 
PKT~\cite{passalisLearningDeepRepresentations2018}, CRD~\cite{tianContrastiveRepresentationDistillation2020}, WCoRD~\cite{chen2021wasserstein}, ReviewKD~\cite{chen2021distilling},
SSKD~\cite{xuKnowledgeDistillationMeets2020a}. 
We directly cite the quantitative results reported in their papers 
~\cite{tianContrastiveRepresentationDistillation2020,chen2021distilling,chen2021wasserstein,olvera2010review}.
For the network of teacher and student models, we use Wide residual networks~\cite{zagoruyko_WideResidualNetworks_2017} (abbreviated as WRNd-w), MobileNetV2~\cite{howard_MobileNetsEfficientConvolutional_2017} (MN2), ShuffleNetV1~\cite{zhang2018shufflenet}\ /ShuffleNetV2~\cite{ma2018shufflenet} (SN1/SN2), and VGG13/VGG8~\cite{simonyan_VeryDeepConvolutional_2015} (V13/V8).
R110, R56 and R20 denote CIFAR-style residual networks, while R50 denotes an ImageNet-style ResNet50.
\textbf{Teacher} and \textbf{Student} stand for the performance of teacher and student models when they are trained individually.

The experimental results on 11 teacher-student pairs are depicted in Tab.\,\ref{tab:SOTA}.
We can see that constructing the proposed knowledge condensation (KC) on vanilla KD shows an impressive improvement.
In addition, our KCD on top of various modern KD approaches all demonstrates an obvious accuracy gain.
More importantly, the proposed KCD utilizes only the condensed knowledge, which 
enjoys the merits of both accuracy and efficiency. 

We also compare KCD with the only existing work that focuses on the computation cost of KD, namely UNIX~\cite{xu_ComputationEfficientKnowledgeDistillation_2020}. 
Tab.\,\ref{tab:UNIX} displays the results on accuracy and computation cost $C$\footnote{Calculation process of $C$ of our method is detailed in appendix.} 
(see Eq.\,(\ref{eq:cost})).
It is noteworthy that the computation $C$ of our KCD is unrelated to the network, thus keeps fixed across different teacher-student pairs.
In contrast, $C$ in UNIX~\cite{xu_ComputationEfficientKnowledgeDistillation_2020} is 
dependent on the rate among the number of sample pass in teacher forward, student forward, and student backward, thus it presents diverse values across different pairs.
KD denotes the vanilla baseline, with $C$ set as $100\%$.
UNIX denotes citing the accuracy results of the models reported in the original works which have the most similar $C$ with our KCD.
UNIX$\dag$ denotes using their public code\footnote{\url{https://github.com/xuguodong03/UNIXKD}} to run and evaluate their methods with the same cost setting $C$ with our method.
It appears that the accuracy of the proposed KCD outperforms UNIX at the same level of computation cost.

\begin{table}[!t] 
  \centering
  \caption{
  Top-1/-5 error (\%) on ImageNet, from ResNet34 to ResNet18. 
  Equipping our method reduces computation cost $C$:100\%$\to$81.61\%, 
  $C_a$:114M$\to$81M. }
  \resizebox{\linewidth}{!}{
    \setlength{\parindent}{0pt}
    \setlength{\tabcolsep}{1.3pt}{
    \begin{tabular}{c|cc|cccc|cc|cc|cc}
    \toprule
    &\multirow{2}{*}{Tea.} & \multirow{2}{*}{Stu.}  &AT    & SP & OnlineKD  &SSKD & KD & KC-KD & CRD& KC-CRD & ReKD& KC-ReKD\\
     &  & & \cite{komodakis2017paying}   & \cite{tungSimilarityPreservingKnowledgeDistillation2019} &  \cite{zhu2018knowledge}  &\cite{xuKnowledgeDistillationMeets2020a} & \cite{hinton2015distilling} &  (Ours)& \cite{tianContrastiveRepresentationDistillation2020} &(Ours) &  \cite{chen2021distilling}& (Ours) \\
    
    \midrule
    Top-1 & 26.69  & 30.25  &29.30  & 29.38 & 29.45&28.38  & 29.34 & 28.61&28.83&28.46 &28.39& {\bf 27.87}\\
    Top-5 & 8.58  & 10.93  &10.00 &10.20& 10.41 & 9.33& 10.12 &9.62 & 9.87& 9.53 & 9.49 & {\bf 9.08}\\
    \bottomrule
    \end{tabular}%
    }
  }
  \label{tab:imagenet}%
\end{table}%

\textbf{Results on ImageNet}.
Following common practice~\cite{tianContrastiveRepresentationDistillation2020,xuKnowledgeDistillationMeets2020a}, 
the experiments on ImageNet are conducted 
using ResNet34 (teacher) and ResNet18 (student).
Tab.\,\ref{tab:imagenet} displays the results of 
both Top-1 and Top-5 error.
We can see that building the proposed knowledge condensation upon KD, CRD and ReviewKD (abbreviated as ReKD) all reduces the testing error significantly.
Moreover, the proposed KC leads to the reduction of relative computation $C$ to $81.61\%$, and the absolute computation $C_a$ from 114M to 81M, which reveals an obvious gain of training efficiency in the large-scale benchmark.

\begin{table}[!t]
  \centering
  \caption{Ablation study of the proposed KCD on four KD processes (\%).}
  \resizebox{0.95\linewidth}{!}{
    \setlength{\parindent}{0pt}
    \setlength{\tabcolsep}{2pt}{
    \begin{tabular}{ccccc}
    \toprule
    Teacher  & {WRN-40-2} & {VGG13} & {VGG13} & {resnet32x4} \\
    Student & {WRN-16-2 } & {VGG8} & {MobileNetV2} & {ShuffleNetV2} \\
    \midrule
    OGVE w/ random  & 74.54  & 73.01  & 67.76  & 74.92 \\
    OGVE w/o OVR   & 75.01  & 73.26  & 67.73  & 74.56  \\
    OGVE w/o CAR  & 75.27  & 73.04  & 67.79  & 74.88  \\
    OGVE -Full  & 75.48  & 73.08  & 68.23  & 75.16  \\
    OGVE + VAKS w/ KA ($\epsilon=\epsilon_m$)    & 75.57  & 73.20  & 68.34  & 75.14  \\
    \midrule
    OGVE + VAKS (Full) & \textbf{75.70}  & \textbf{73.44}  & \textbf{68.61}  & \textbf{75.19}  \\
    \bottomrule
    \end{tabular}%
  }
  }
  \label{tab:ablation}%
\end{table}%

\subsection{Further Empirical Analysis}
\label{s2}
\textbf{Ablation Study}.
We verify the effect of each component in the proposed framework by conducting ablation studies. 
The results are provided in Tab.\,\ref{tab:ablation}.
\textbf{(1)} OGVE w/ random denotes we randomly allocate ranking position $\mathcal{R}_V$ as well as value label $Y$ instead of using OGVE.
\textbf{(2)} OGVE w/o OVR, w/o CAR, -full denotes that we remove online value recording (\emph{i.e.}, estimate the value during the mini-batch training), cost-aware ranking (\emph{i.e.}, discard the weight $\big(F(x)\big)^\alpha$ in Eq.\,(\ref{eq:rank}), and keep the full setting of OGVE.
Note that the above variants of OGVE are combined with a direct selection on label $y$=$1$.
\textbf{(3)} OGVE + VAKS w/ KA ($\epsilon$=$\epsilon_m$) denotes using non-adaptive knowledge augmentation with $\epsilon$ keeping its maximum
$\epsilon_m$=$0.3$ 
for all points.
OGVE + VAKS denotes the full structure of our proposed KCD.
When any component is removed, it appears that the performance drops accordingly, revealing the effectiveness of our design.

\begin{figure}[!t]  
  \centering
  \subfloat[Accuracy with ratio $\rho$] 
  {
      \begin{minipage}[t]{0.35\linewidth}
          \centering      
          \includegraphics[width=\linewidth,height=2.8cm]{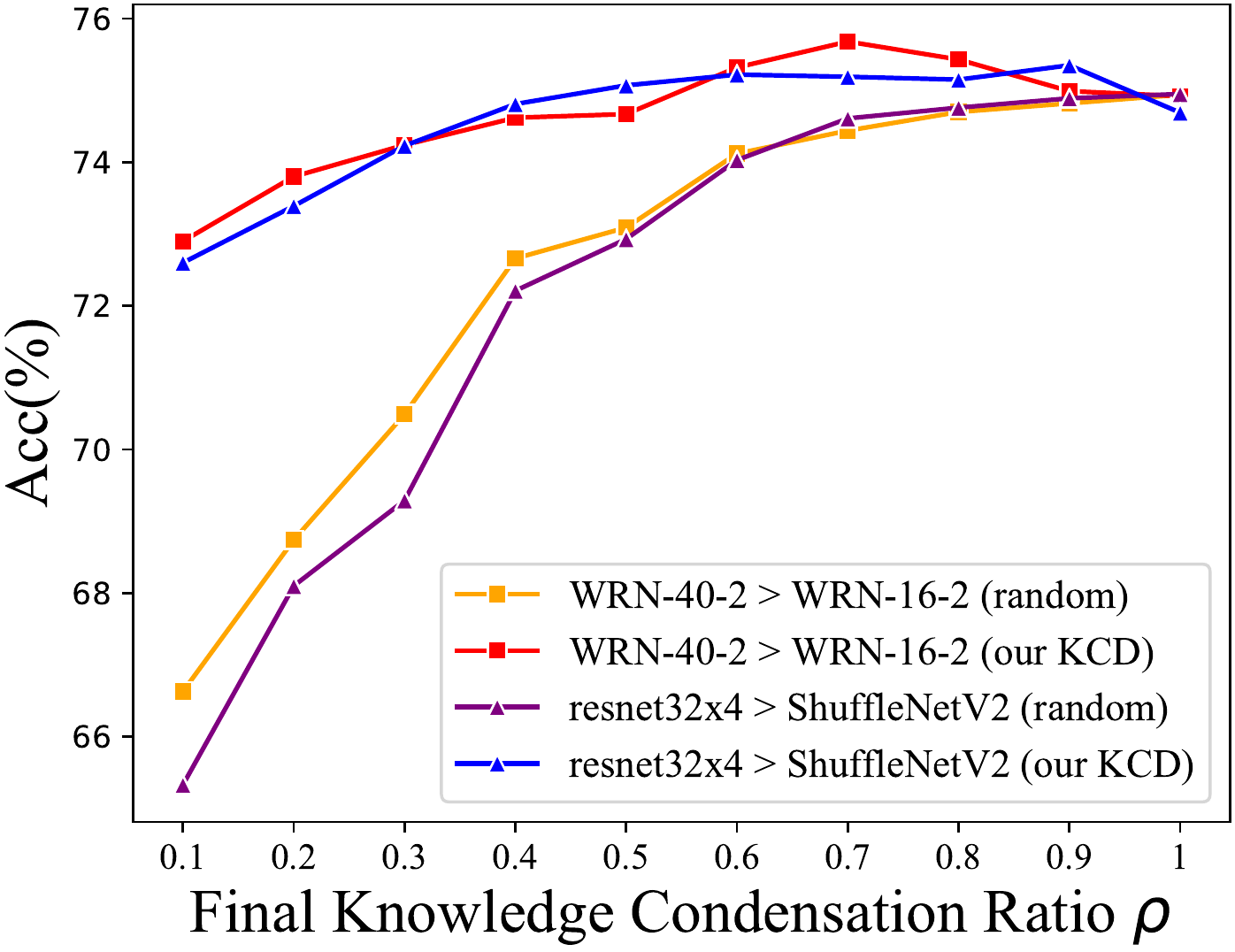} 
          \label{fig:r}
      \end{minipage}
  }%
   \subfloat[Varying knowledge value] 
  {
      \begin{minipage}[t]{0.32\linewidth}
          \centering          
          \includegraphics[width=\linewidth,height=2.9cm]{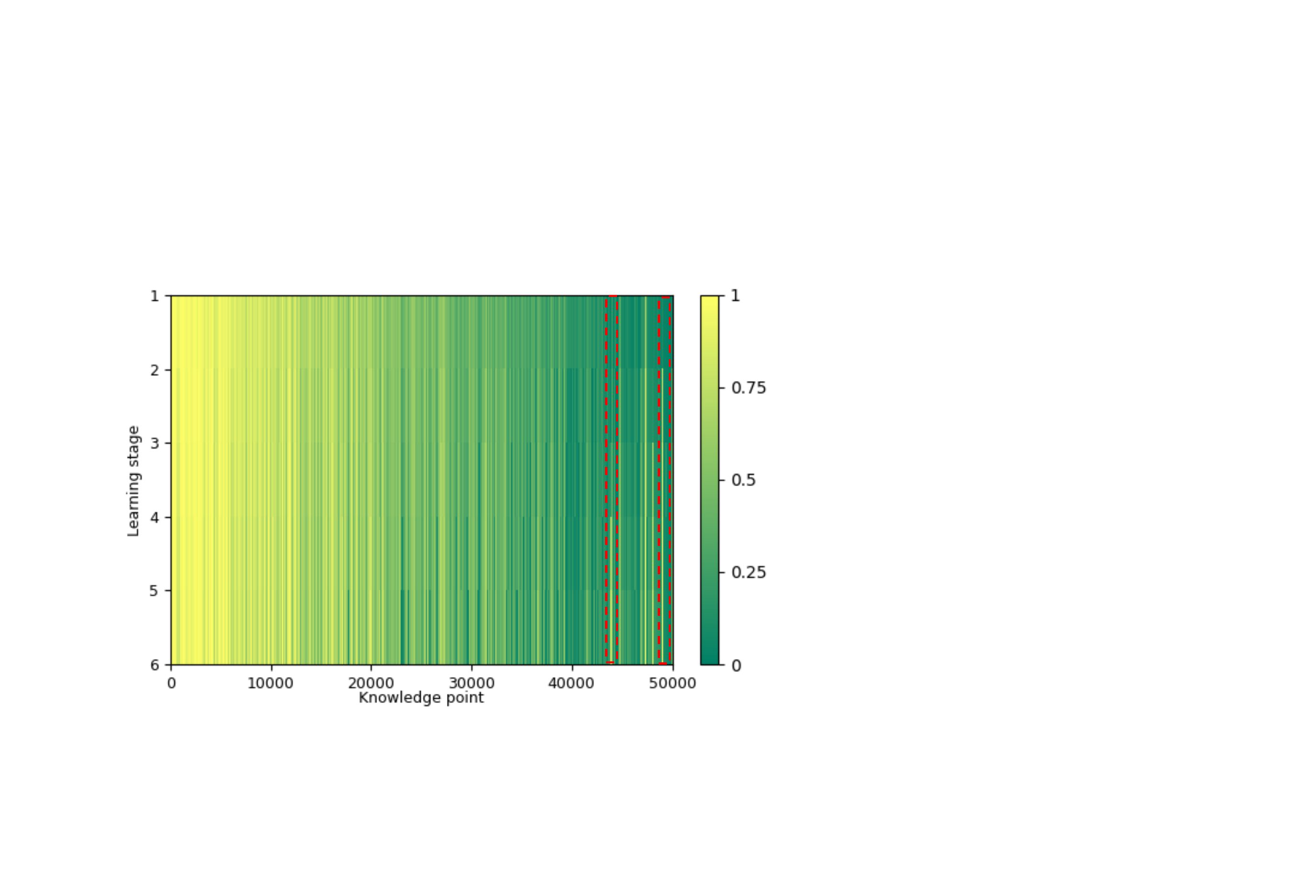}   
          \label{fig:transfer}
      \end{minipage}%
  }
  \subfloat[Distance matrix] 
  {
      \begin{minipage}[t]{0.3\linewidth}
          \centering          
          \includegraphics[width=\linewidth,height=2.8cm]{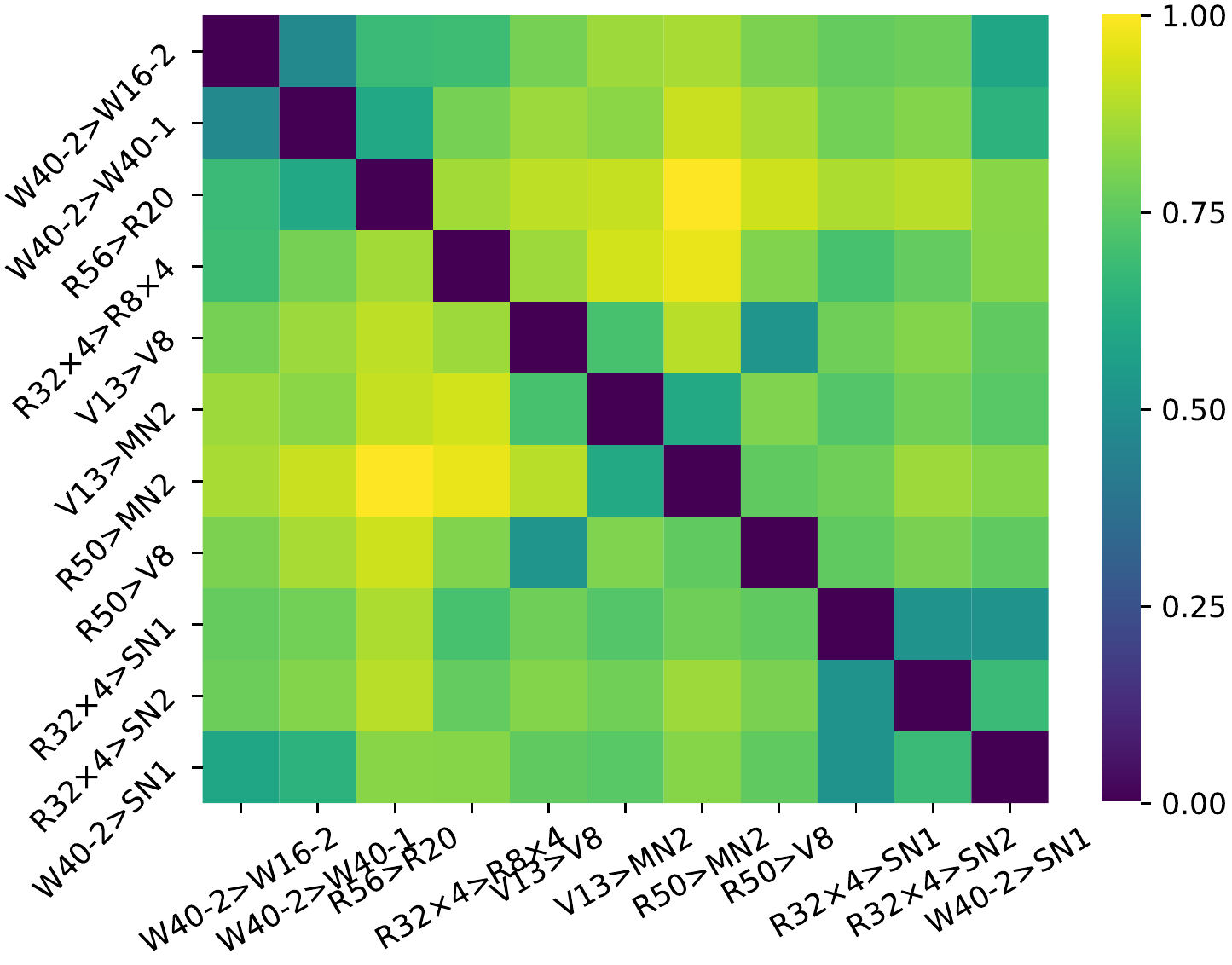}  
          \label{fig:cm}
      \end{minipage}%
  } 
  \caption{
  (a) Accuracy with the variation of condensation ratio $\rho$. 
  (b) Pattern of varying knowledge value across the training process.
  (c) Hamming Distance matrix of the value label across KD processes. 
  } 
  \label{fig:further}  
\end{figure}

\textbf{Influence of Knowledge Condensation Ratio  \boldsymbol{$\rho$}}.
Fig.\,\ref{fig:further}(a) displays the performance of different models (\emph{i.e.}, random selection baseline and our KCD) on different KD processes (\emph{i.e.}, 
W40-2$>$W16-2 and R32$\times$4$>$SN2), with the variation of final knowledge condensation ratio $\rho$.
We can see that our KCD outperforms the random baseline by a significant margin, especially as $\rho$ decreasing. 
It is noteworthy that the proposed KCD achieves better results on $\rho$ ranging between 0.6-0.8 than full-knowledge setting with $\rho$=1, and nearly maintains the accuracy among a wide range $\rho$ as 0.3-1.0, which implies that our KCD can identify and summarize the compact yet effective knowledge encoding that is robust to the size reduction of knowledge set.

\textbf{Pattern of Knowledge Value}.
Fig.\,\ref{fig:further}(b) displays the varying ranking-based probability {\emph{w.r.t}} knowledge value during the entire training process.
We can see that the value of knowledge points differs to the student at different learning stages.
As marked in red, some knowledge points are valueless at the beginning stage while become more and more critical in the later stages.
Fig.\,\ref{fig:further}(c) depicts the Hamming distance matrix about the estimated value label $Y$ in the final stage across various KD processes, wherein the distance indicates the number of different elements in two masks. 
We can see that the value label 
represents 
a relatively strong correlation (small distance) when two KD processes 
have the same student architectures
(\emph{e.g.}, V13$>$V8 and R50$>$V8) 
or similar ones (\emph{e.g.}, R32$\times$4$>$SN1 and R32$\times$4$>$SN2),
revealing that the identified knowledge value really encodes some ``patterns'' of the student model.

\begin{figure}[!h]   
	\centering	   
	\includegraphics[width=0.95\linewidth]{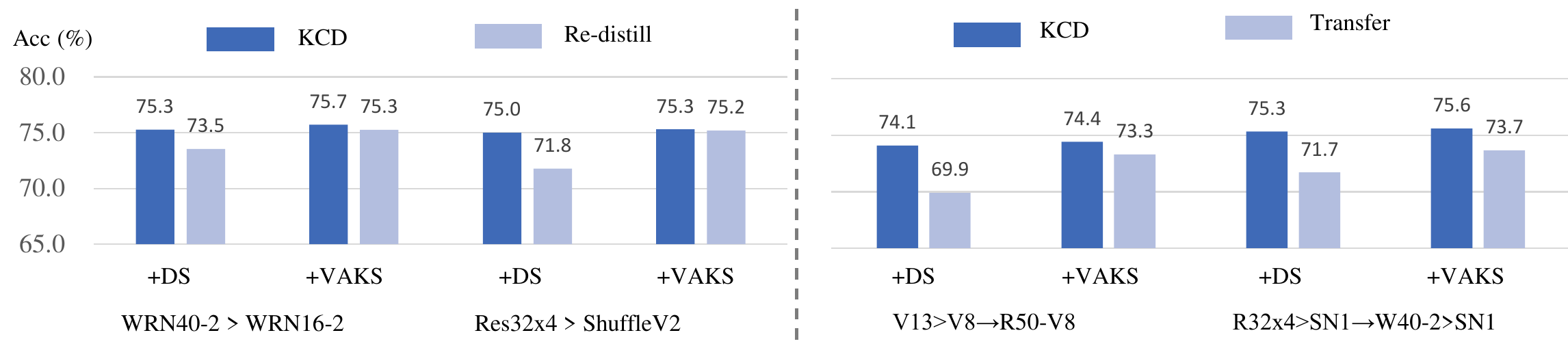}
	\caption{
	The performance of reusing the condensed knowledge. \textbf{Left}:
	We utilize the condensed knowledge to directly re-train the student model in the original KD process from scratch. \textbf{Right}: We transfer the condensed knowledge encoding to facilitate another KD process.
	} \label{fig:reuse}   
\end{figure} 

\textbf{Reuse of Condensed Knowledge}.
The observed similarity of knowledge value across KD processes 
inspires us to investigate on reusing the condensed knowledge for efficient training.
As shown in Fig.\,\ref{fig:reuse}(\textbf{Left}), 
we first use the ready-made condensed knowledge to re-distill the student from a scratch.
``+DS'' (direct selection) and ``+VAKS'' (value-adaptive knowledge summary) denote two variants of our KCD.
It appears that compared with our standard KCD, the performance of the re-distilled student drops significantly when equipped with ``+DS'' while achieves comparable results with ``+VAKS''.
As shown in Fig.\,\ref{fig:reuse}(\textbf{Right}), 
we further evaluate the transferability of the knowledge condensation, where 
we transfer the knowledge encoding condensed in a source KD process to a target KD process to improve the efficiency.
As can be seen, the performance of transferring the condensed knowledge degrades dramatically compared with the standard KCD.
In comparison, when equipping the transfer process with our VASK module, the performance gap with standard KCD is reduced a lot.
These observations demonstrate the potential of our KCD method for promoting the efficient training by reusing and transferring the condensed knowledge.

\section{Conclusion}
This paper proposes Knowledge Condensation Distillation (KCD) to address the knowledge redundancy during KD.
Instead of relying on the whole knowledge set from teacher model, the key idea is to first identify the informative knowledge components and then summarize a compact knowledge encoding to perform KD efficiently.
Specially, we forge an iterative optimization framework between condensing the compact knowledge encoding and compressing the student model based on the EM algorithm.
We further present two collaborative modules to perform the proposed KCD, as
online global value estimation (OGVE) and value-adaptive knowledge summary (VAKS).
Extensive experiments demonstrate the effectiveness of the proposed KCD against the state-of-the-arts.

\noindent\textbf{Acknowledgement.}
The study is supported partly by the National Natural Science Foundation of China under Grants 82172033, U19B2031, 61971369, 52105126, China, in part of Science and Technology Key Project of Fujian Province(No. 2019HZ020009).



\clearpage
%
%
\bibliographystyle{splncs04}
\bibliography{reference}

\begin{thebibliography}{10}
\providecommand{\url}[1]{\texttt{#1}}
\providecommand{\urlprefix}{URL }
\providecommand{\doi}[1]{https://doi.org/#1}

\bibitem{ahnVariationalInformationDistillation2019a}
Ahn, S., Hu, S.X., Damianou, A., Lawrence, N.D., Dai, Z.: Variational
  information distillation for knowledge transfer. In: Proceedings of the IEEE
  Conference on Computer Vision and Pattern Recognition (CVPR). pp. 9163--9171
  (2019)

\bibitem{caiProxylessNASDirectNeural2019}
Cai, H., Zhu, L., Han, S.: Proxylessnas: Direct neural architecture search on
  target task and hardware. In: Proceedings of the International Conference of
  Learning Representation (ICLR) (2019)

\bibitem{chen2021wasserstein}
Chen, L., Wang, D., Gan, Z., Liu, J., Henao, R., Carin, L.: Wasserstein
  contrastive representation distillation. In: Proceedings of the IEEE
  Conference on Computer Vision and Pattern Recognition (CVPR). pp.
  16296--16305 (2021)

\bibitem{chen2021distilling}
Chen, P., Liu, S., Zhao, H., Jia, J.: Distilling knowledge via knowledge
  review. In: Proceedings of the IEEE Conference on Computer Vision and Pattern
  Recognition (CVPR). pp. 5008--5017 (2021)

\bibitem{IMAGENET}
Deng, J., Dong, W., Socher, R., Li, L.J., Li, K., Fei-Fei, L.: Imagenet: A
  large-scale hierarchical image database. In: Proceedings of the IEEE
  Conference on Computer Vision and Pattern Recognition (CVPR). pp. 248--255
  (2009)

\bibitem{goodfellow2014explaining}
Goodfellow, I.J., Shlens, J., Szegedy, C.: Explaining and harnessing
  adversarial examples. arXiv preprint arXiv:1412.6572  (2014)

\bibitem{han2018co}
Han, B., Yao, Q., Yu, X., Niu, G., Xu, M., Hu, W., Tsang, I., Sugiyama, M.:
  Co-teaching: Robust training of deep neural networks with extremely noisy
  labels. Proceedings of the Advances in Neural Information Processing Systems
  (NeurIPS)  \textbf{31} (2018)

\bibitem{har2007smaller}
Har-Peled, S., Kushal, A.: Smaller coresets for k-median and k-means
  clustering. Discrete \& Computational Geometry  \textbf{37}(1),  3--19 (2007)

\bibitem{heoComprehensiveOverhaulFeature2019}
Heo, B., Kim, J., Yun, S., Park, H., Kwak, N., Choi, J.Y.: A comprehensive
  overhaul of feature distillation. In: Proceedings of the IEEE Conference on
  Computer Vision and Pattern Recognition (CVPR). pp. 1921--1930 (2019)

\bibitem{hinton2015distilling}
Hinton, G., Vinyals, O., Dean, J.: Distilling the knowledge in a neural
  network. arXiv preprint arXiv:1503.02531  (2015)

\bibitem{howard_MobileNetsEfficientConvolutional_2017}
Howard, A.G., Zhu, M., Chen, B., Kalenichenko, D., Wang, W., Weyand, T.,
  Andreetto, M., Adam, H.: Mobilenets: Efficient convolutional neural networks
  for mobile vision applications. arXiv preprint arXiv:1704.04861  (2017)

\bibitem{katharopoulos2018not}
Katharopoulos, A., Fleuret, F.: Not all samples are created equal: Deep
  learning with importance sampling. In: Proceedings of the International
  Conference on Machine Learning (ICML). pp. 2525--2534 (2018)

\bibitem{komodakis2017paying}
Komodakis, N., Zagoruyko, S.: Paying more attention to attention: improving the
  performance of convolutional neural networks via attention transfer. In:
  Proceedings of the International Conference of Learning Representation (ICLR)
  (2017)

\bibitem{CIFAR100}
Krizhevsky, A., Hinton, G., et~al.: Learning multiple layers of features from
  tiny images  (2009)

\bibitem{li2020prototypical}
Li, J., Zhou, P., Xiong, C., Hoi, S.C.: Prototypical contrastive learning of
  unsupervised representations. In: Proceedings of the International Conference
  of Learning Representation (ICLR

\bibitem{li2022learning}
Li, S., Lin, M., Wang, Y., Fei, C., Shao, L., Ji, R.: Learning efficient gans
  for image translation via differentiable masks and co-attention distillation.
  IEEE Transactions on Multimedia (TMM)  (2022)

\bibitem{li2022distilling}
Li, S., Lin, M., Wang, Y., Wu, Y., Tian, Y., Shao, L., Ji, R.: Distilling a
  powerful student model via online knowledge distillation. IEEE Transactions
  on Neural Networks and Learning Systems (TNNLS)  (2022)

\bibitem{linHRankFilterPruning2020a}
Lin, M., Ji, R., Wang, Y., Zhang, Y., Zhang, B., Tian, Y., Shao, L.: Hrank:
  Filter pruning using high-rank feature map. In: Proceedings of the IEEE
  Conference on Computer Vision and Pattern Recognition (CVPR). pp. 1529--1538
  (2020)

\bibitem{ma2018shufflenet}
Ma, N., Zhang, X., Zheng, H.T., Sun, J.: Shufflenet v2: Practical guidelines
  for efficient cnn architecture design. In: Proceedings of the European
  Conference on Computer Vision (ECCV). pp. 116--131 (2018)

\bibitem{mirzasoleiman2020coresets}
Mirzasoleiman, B., Bilmes, J., Leskovec, J.: Coresets for data-efficient
  training of machine learning models. In: Proceedings of the International
  Conference on Machine Learning (ICML). pp. 6950--6960 (2020)

\bibitem{muller_WhenDoesLabel_2020}
M{\"u}ller, R., Kornblith, S., Hinton, G.E.: When does label smoothing help?
  Proceedings of the Advances in Neural Information Processing Systems
  (NeurIPS)  (2019)

\bibitem{olvera2010review}
Olvera-L{\'o}pez, J.A., Carrasco-Ochoa, J.A., Mart{\'\i}nez-Trinidad, J.F.,
  Kittler, J.: A review of instance selection methods. Artificial Intelligence
  Review  \textbf{34}(2),  133--143 (2010)

\bibitem{parkRelationalKnowledgeDistillation2019}
Park, W., Kim, D., Lu, Y., Cho, M.: Relational knowledge distillation. In:
  Proceedings of the IEEE Conference on Computer Vision and Pattern Recognition
  (CVPR). pp. 3967--3976 (2019)

\bibitem{passalisLearningDeepRepresentations2018}
Passalis, N., Tefas, A.: Learning deep representations with probabilistic
  knowledge transfer. In: Proceedings of the European Conference on Computer
  Vision (ECCV). pp. 268--284 (2018)

\bibitem{romero2014fitnets}
Romero, A., Ballas, N., Kahou, S.E., Chassang, A., Gatta, C., Bengio, Y.:
  Fitnets: Hints for thin deep nets. arXiv preprint arXiv:1412.6550  (2014)

\bibitem{sener2017active}
Sener, O., Savarese, S.: Active learning for convolutional neural networks: A
  core-set approach. arXiv preprint arXiv:1708.00489  (2017)

\bibitem{shen_LabelSmoothingTruly_2021}
Shen, Z., Liu, Z., Xu, D., Chen, Z., Cheng, K.T., Savvides, M.: Is label
  smoothing truly incompatible with knowledge distillation: An empirical study.
  In: Proceedings of the International Conference of Learning Representation
  (ICLR) (2020)

\bibitem{simonyan_VeryDeepConvolutional_2015}
Simonyan, K., Zisserman, A.: Very deep convolutional networks for large-scale
  image recognition. arXiv preprint arXiv:1409.1556  (2014)

\bibitem{szegedy2013intriguing}
Szegedy, C., Zaremba, W., Sutskever, I., Bruna, J., Erhan, D., Goodfellow, I.,
  Fergus, R.: Intriguing properties of neural networks. arXiv preprint
  arXiv:1312.6199  (2013)

\bibitem{tianContrastiveRepresentationDistillation2020}
Tian, Y., Krishnan, D., Isola, P.: Contrastive representation distillation. In:
  Proceedings of the International Conference of Learning Representation (ICLR)
  (2019)

\bibitem{toneva2018empirical}
Toneva, M., Sordoni, A., des Combes, R.T., Trischler, A., Bengio, Y., Gordon,
  G.J.: An empirical study of example forgetting during deep neural network
  learning. In: Proceedings of the International Conference of Learning
  Representation (ICLR) (2018)

\bibitem{tungSimilarityPreservingKnowledgeDistillation2019}
Tung, F., Mori, G.: Similarity-preserving knowledge distillation. In:
  Proceedings of the IEEE International Conference on Computer Vision (ICCV).
  pp. 1365--1374 (2019)

\bibitem{wang2018dataset}
Wang, T., Zhu, J.Y., Torralba, A., Efros, A.A.: Dataset distillation. arXiv
  preprint arXiv:1811.10959  (2018)

\bibitem{xuKnowledgeDistillationMeets2020a}
Xu, G., Liu, Z., Li, X., Loy, C.C.: Knowledge distillation meets
  self-supervision. In: Proceedings of the European Conference on Computer
  Vision (ECCV). pp. 588--604 (2020)

\bibitem{xu_ComputationEfficientKnowledgeDistillation_2020}
Xu, G., Liu, Z., Loy, C.C.: Computation-efficient knowledge distillation via
  uncertainty-aware mixup. arXiv preprint arXiv:2012.09413  (2020)

\bibitem{yamamotoLearnableCompandingQuantization2021}
Yamamoto, K.: Learnable companding quantization for accurate low-bit neural
  networks. In: Proceedings of the IEEE Conference on Computer Vision and
  Pattern Recognition (CVPR). pp. 5029--5038 (2021)

\bibitem{yin2020dreaming}
Yin, H., Molchanov, P., Alvarez, J.M., Li, Z., Mallya, A., Hoiem, D., Jha,
  N.K., Kautz, J.: Dreaming to distill: Data-free knowledge transfer via
  deepinversion. In: Proceedings of the IEEE Conference on Computer Vision and
  Pattern Recognition (CVPR) (2020)

\bibitem{zagoruyko_WideResidualNetworks_2017}
Zagoruyko, S., Komodakis, N.: Wide residual networks. arXiv preprint
  arXiv:1605.07146  (2016)

\bibitem{zhang2018shufflenet}
Zhang, X., Zhou, X., Lin, M., Sun, J.: Shufflenet: An extremely efficient
  convolutional neural network for mobile devices. In: Proceedings of the IEEE
  Conference on Computer Vision and Pattern Recognition (CVPR). pp. 6848--6856
  (2018)

\bibitem{zhang2021efficient}
Zhang, Z., Chen, X., Chen, T., Wang, Z.: Efficient lottery ticket finding: Less
  data is more. In: Proceedings of the International Conference on Machine
  Learning (ICML). pp. 12380--12390 (2021)

\bibitem{zhao2020dataset}
Zhao, B., Mopuri, K.R., Bilen, H.: Dataset condensation with gradient matching.
  In: Proceedings of the International Conference of Learning Representation
  (ICLR) (2020)

\bibitem{zhu2018knowledge}
Zhu, X., Gong, S., et~al.: Knowledge distillation by on-the-fly native
  ensemble. Proceedings of the Advances in Neural Information Processing
  Systems (NeurIPS)  (2018)

\end{thebibliography}

\section*{Appendix}

\section*{Calculation of Computation Cost}


In what follows, we first shortly describe the metrics of computation cost in UNIX~\cite{xu_ComputationEfficientKnowledgeDistillation_2020}
and then show that they are equivalent to the metrics of cost $C$ used in this paper (Sec.\,\ref{sec:efficiency}).

\textbf{Computation Metrics in UNIX}.
Considering the sampling number and the computation in different network passes, the cost of computation
in UNIX
is calculated by:
\begin{equation}
  \label{eq2}
  \begin{aligned}
  E & =  N_t \cdot F_t  +  N_{s1} \cdot F_s + N_{s2} \cdot B_s,
  \end{aligned}
\end{equation}
where $F_t$, $F_s$ and $B_s$ denote the float-point operation number in teacher forward pass, student forward pass and student backward pass while $N_t$, $N_{s1}$, $N_{s2}$ denote their total sampling number over the entire training procedure.

For a vanilla KD baseline, the sampling number in different passes keeps fixed, \emph{i.e.}, $N_t=N_{s1}=N_{s2}$.
By denoting this value as $N$, the baseline cost can be derived as: $E=N\cdot( F_t  +  F_s + B_s)$. 
As a comparison, for UNIX, $N_{t}$ and $N_{s2}$ are reduced to $N_k$ (where $N_k<N$) while $N_{s1}$ is increased to $N+N_k$, which makes $E=N_k \cdot F_t  +  (N+N_k) \cdot F_s + N_k \cdot B_s$.
$Computation$ is calculated by the ratio of them:
\begin{equation}
  \label{eq3}
  \begin{aligned}
  Computation &=  \frac{N_k \cdot F_t  +  (N+N_k) \cdot F_s + N_k \cdot B_s}{N \cdot F_t  +  N \cdot F_s + N \cdot B_s}\\
  &=  \frac{N_k \cdot 1  +  (N+N_k) \cdot \frac{F_s}{F_t} + N_k \cdot \frac{B_s}{F_t}}{N \cdot 1  +  N \cdot \frac{F_s}{F_t} + N \cdot \frac{B_s}{F_t}}. \
  \end{aligned}
\end{equation}

In UNIX~\cite{xu_ComputationEfficientKnowledgeDistillation_2020}, the approximation of $B_s\approx F_s$ is introduced, so Eq.\,(\ref{eq3}) can be re-written as: 
\begin{equation}
  \label{eq4}
  \begin{aligned}
  Computation 
  \approx \frac{N_k \cdot 1  +  (N+2N_k) \cdot \frac{F_s}{F_t}}{N \cdot 1  +  2N \cdot \frac{F_s}{F_t}}.
  \end{aligned}
\end{equation}
where $\frac{F_s}{F_t}$ denotes the ratio of float-point operation number between student and teacher forward passes, which varies in different teacher-student pairs.

\textbf{Relation of Computation Metrics Between This Paper and UNIX}.
In this paper, the absolute computation cost $C_a$ counts the total sampling number over the training procedure.
Formally, given $I$ training epochs, $T$ epochs in a condensation stage, condensation threshold$\tau^t$ at the $t$-th stage  
$C_a=|K|\cdot I$ provides the computation for the vanilla KD baseline.
For our KCD, $C_a = |K|\cdot (\tau^0 + \cdots + \tau^t + \cdots + \tau^{I/T})\cdot T$.
By calculating the ratio, the relative computation cost $C$ can be obtained, as shown in Eq.\,(\ref{eq:cost}).

In fact, if we analyze the computation of our KCD by Eq.\,(\ref{eq2}), we have:
\begin{equation}
  \label{eq5}
  E_{KCD} = |K|\cdot (\tau^0 + \cdots + \tau^t + \cdots + \tau^{I/T})\cdot T \cdot(F_t+F_s+B_s). 
\end{equation}

Further considering the vanilla KD baseline as $E=|K|\cdot I \cdot (F_t+F_s+B_s)$, the $Computation$ for our KCD can be calculated as:
\begin{equation}
  \label{eq6}
  \begin{aligned}
  Computation_{KCD} &=  \frac{|K|\cdot (\tau^0 + \cdots + \tau^t + \cdots + \tau^{I/T})\cdot T \cancel{\cdot (F_t+F_s+B_s)}}{|K|\cdot I 
  \cancel{\cdot (F_t+F_s+B_s)}} \\
  &= C.
  \end{aligned}
\end{equation}

Thus the metric of $computation$ in UNIX is equivalent to $C$ in Eq.\,(16) of the main paper.
This means that the computation results of our KCD are comparable with those in UNIX, as shown in Tab.\,\ref{tab:UNIX} and Tab.\,\ref{tab:imagenet} of our main paper.
It is noteworthy that the $computation$ or $C$ for our KCD is only dependent on the compactness of knowledge encoding, thus keeps unchanged across different teacher-student pairs.

\textbf{Results in Tab.\,\ref{tab:UNIX} of Our Main Paper}.
In Tab.\,\ref{tab:UNIX} of our main paper, $C=100\%$ is set for the vanilla KD baseline while $C=81.6\%$ is calculated via the derivation of Eq.\,(\ref{eq:cost}) in our main paper, as $C = \frac{\rho^\frac{I}{T}(1-\rho)}{1-\rho^\frac{T}{T}}\cdot\frac{T}{I}$ (derived from the exponential decaying of the condensation threshold $\tau$). 
For the second row, we directly cite the results in UNIX with the most similar $computation$ or $C$ to ours.
For the third row, we run the official code of UNIX
with the adjusted parameters of $N_k$ in Eq.\,(\ref{eq4}) to make $Computation$ be equal to our KCD.
We can observe that the accuracy of the proposed KCD
outperforms UNIX at the same level of computation cost.

\textbf{Results in Tab.\,\ref{tab:imagenet} of Our Main Paper}.
$C=81.6\%$ on ImageNet is calculated in the same way with CIFAR100 in Tab.\,\ref{tab:UNIX} of our main paper.
$C_a$ of vanilla KD baseline is calculated via $1.26M\times90=114M$.
$C_a$ of the proposed KCD is calculated by $114M\times C=81M$.
As can be seen, our KCD reveals an obvious gain of distillation accuracy and efficiency on the large-scale benchmark.

\section*{More Ablation Studies}
\textbf{Influence of Cost-Aware Weighting Coefficient \boldsymbol{$\alpha$}}.
We evaluate the sensitivity of the proposed KCD \emph{w.r.t.} \boldsymbol{$\alpha$}
 (in Eq. (\ref{eq:rank}) of our main paper) in three KD processes on CIFAR100.
The results are reported in Tab.\,\ref{tab1}.
We vary the parameter in \{0.01, 0.03, 0.05, 0.1\}, and choose \boldsymbol{$\alpha=0.03$} for its best performance.

\begin{table}[!h]
  \centering
  \caption{Ablation study of $\alpha$ on CIFAR100 (Acc\%).}
     \resizebox{0.55\linewidth}{!}{
    \setlength{\parindent}{0pt}
    \setlength{\tabcolsep}{3pt}{
    \begin{tabular}{cccc}
    \toprule
    {Teacher} & {wrn-40-2} & {vgg13} & {resnet32x4} \\
    {Student} & {wrn-16-2 } & {mobilenetv2} & {ShuffleNetV2} \\
    \midrule
    0.01  & 75.27 & 66.40  & 74.85 \\
    0.03  & \textbf{75.70}  & \textbf{68.61} & 75.19 \\
    0.05  & 74.99 & 67.86 & \textbf{75.23} \\
    0.1   & 75.43 & 68.28 & 75.11 \\
    \bottomrule
    \end{tabular}%
    }
    }
  \label{tab1}%
\end{table}%

\begin{table}[!h]
  \centering
  \caption{Ablation study of $\rho$ on CIFAR100 (Acc\%).}
    \resizebox{0.55\linewidth}{!}{
    \setlength{\parindent}{0pt}
    \setlength{\tabcolsep}{3pt}{
    \begin{tabular}{cccc}
    \toprule
    Teacher  & {wrn-40-2}  &{vgg13} &{resnet32x4}   \\
    Student & {wrn-16-2 } & {mobilenetv2}  & {ShuffleNetV2} \\
    \midrule
    0.9 & 75.01  & 68.23 & \textbf{75.36}    \\
    0.7 & \textbf{75.70}  & \textbf{68.61}& 75.19     \\
    0.5 & 74.89  & 66.81 & 75.08     \\
    0.3   & 74.07  & 65.50& 74.03     \\
    \midrule
    \end{tabular}%
  }
  }
  \label{tab2}%
\end{table}%

\textbf{Influence of Final Condensation Ratio \boldsymbol{$\rho$}}.
We further evaluate the sensitivity of the proposed KCD \emph{w.r.t.} final condensation ratio of knowledge encoding, \boldsymbol{$\rho$} (in Alg.\,{\ref{alg}} of our main paper) on CIFAR100.
The results are reported in Tab.\,\ref{tab2}.
We vary the parameter in \{0.9, 0.7, 0.5, 0.3\}, and choose \boldsymbol{$\rho=0.7$} due to its best performance.

\begin{table}[!t]
  \centering
  \caption{Ablation study  on CIFAR100 (Acc\%).}
    \resizebox{0.75\linewidth}{!}{
    \setlength{\parindent}{0pt}
    \setlength{\tabcolsep}{3pt}{
    \begin{tabular}{cccc}
    \toprule
    Teacher  & {wrn-40-2}  &{vgg13} &{resnet32x4}   \\
    Student & {wrn-16-2 } & {mobilenetv2}  & {ShuffleNetV2} \\
    \midrule
    $K_1$ w/o Aug. & 75.45  & 68.23 & 75.16 \\
    Aug. $K_1$ in random ($\epsilon=\epsilon_m$) & 75.41  & 65.94 & 75.12   \\
    Aug. $K_{1H}$ & 75.52  & 68.02 & 74.77    \\
    Aug. $K_{1L}$ & \textbf{75.70}  & \textbf{68.61} & \textbf{75.19}     \\
    \midrule
    \end{tabular}%
  }
  }
  \label{tab3}%
\end{table}%

\textbf{Design of Knowledge Augmentation}.
Further, we evaluate the performance of the proposed value-adaptive knowledge summary (VAKS) module equipped with different knowledge augmentation strategies on CIFAR100.
The results are displayed in Tab.\,\ref{tab3}.
\boldsymbol{$K_1$} \textbf{w/o Aug.} denotes that we remove the knowledge augmentation in our work, letting $\hat K=K_1$ in our main paper. 
\textbf{Aug.} \boldsymbol{$K_1$} \textbf{in random} denotes that we utilize $K_0$ to randomly augment $|K_{0}|$ knowledge points in $K_1$ with $\epsilon=\epsilon_m$ (as in Eq.\,(14) of our main paper).
\textbf{Aug.} \boldsymbol{$K_{1H}$} denotes that we re-partition $K_1$ to $K_{1H}$ and $K_{1L}$ so that $|K_{1H}|=|K_0|$, and augment $K_{1H}$.
\textbf{Aug.} \boldsymbol{$K_{1L}$} corresponds to the final design in our main paper, which achieves the best performance.

\begin{figure}[!t]  
  \centering    
 
  \subfloat[Images of valuable or valueless knowledge] 
  {
      \begin{minipage}[t]{0.5\linewidth}
          \centering      
          \includegraphics[width=\linewidth]{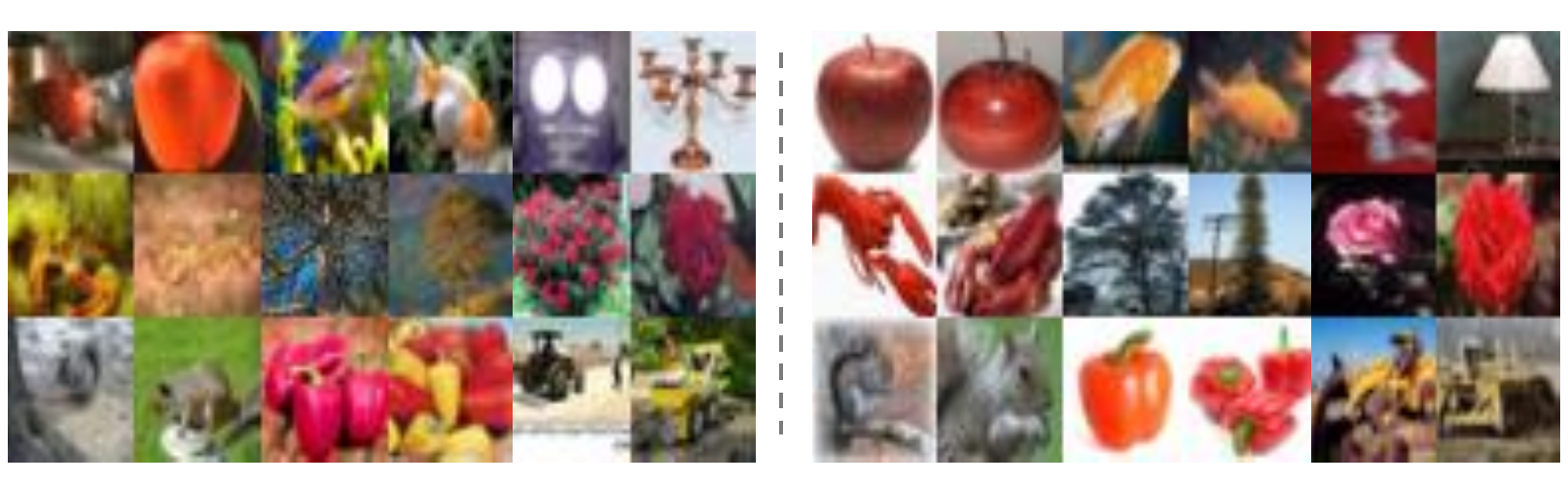}  
          \label{fig:in}
      \end{minipage}
  }%
  \subfloat[Average entropy of soft labels] 
  {
      \begin{minipage}[t]{0.4\linewidth}
          \centering          
          \includegraphics[width=\linewidth]{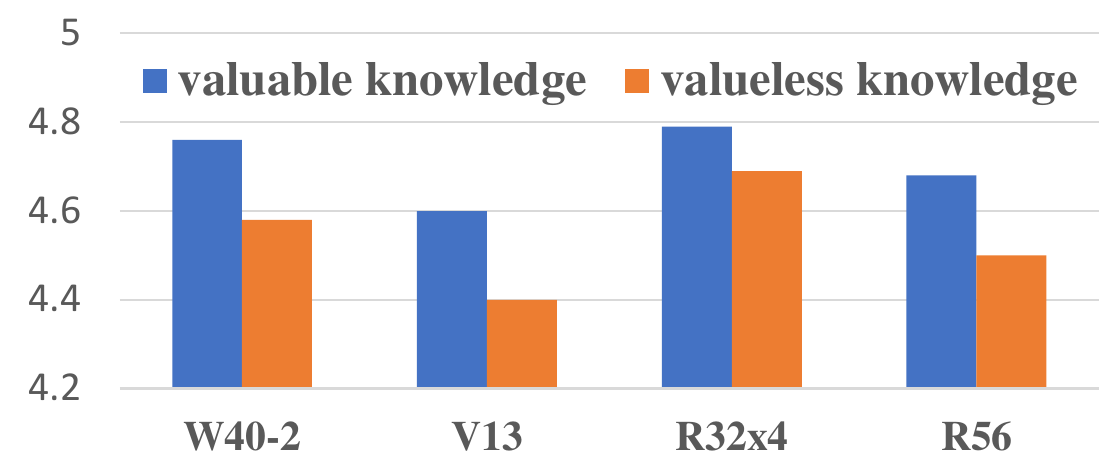} 
          \label{fig:hist}
      \end{minipage}%
  }
  \caption{
Visualization of images and knowledge hints in knowledge points with value label $y=1$ or 0.
  } 
  \label{fig:pattern}  
\end{figure}

\section*{Visualization of Valuable Knowledge Points}
Fig.\,\ref{fig:pattern} displays the visualization of the knowledge points with value label $y=1$ or 0, which is conducted on CIFAR100.
As shown in Fig.\,\ref{fig:pattern}(a)(\textbf{Left}), 
the images from ``valueless'' knowledge show less complexity where objects are centered and easily distinguishable.
In comparison, the ``valuable'' images shown in Fig.\,\ref{fig:pattern}(a)(\textbf{Right}) tend to contain multiple ambiguous elements that are lower-quality with complicated backgrounds and more challenging to recognize.
Fig.\,\ref{fig:pattern}(b) reveals the ``valuable'' patterns of knowledge hints from the average entropy of soft labels via four teacher models.
It appears that the soft labels of valuable knowledge points tend to have a higher entropy, implying more informative semantic structural information.

\end{document}